\documentclass[letterpaper]{article}
\usepackage[margin=2.5cm]{geometry}
\usepackage{times} \usepackage{helvet} \usepackage{courier} \usepackage[hyphens]{url} \usepackage{graphicx} 
\usepackage{caption}
\usepackage{authblk}
\DeclareCaptionStyle{ruled}{labelfont=normalfont,labelsep=colon,strut=off}    \usepackage{subcaption}

\title{Planning with Biological Neurons and Synapses}
\author[1]{Francesco d'Amore}
\author[2]{Daniel Mitropolsky}
\author[3]{Pierluigi Crescenzi}
\author[1]{Emanuele Natale}
\author[2]{Christos H.\ Papadimitriou}
\affil[1]{Universit\'e C\^ote d'Azur, Inria, CNRS, I3S, Sophia Antipolis, France }
\affil[2]{Department of Computer Science, Columbia University, New York, USA }
\affil[3]{Gran Sasso Science Institute, L'Aquila, Italia {\protect\\ \tt \{francesco.d-amore,emanuele.natale\}@inria.fr}{\protect\\ \tt \{dgm2144,christos\}@columbia.edu}{\protect\\ \tt pierluigi.crescenzi@gssi.it}}

\newcommand{\projectstar}{\mathtt{strongProject}()}
\newcommand{\project}[2]{\mathtt{project}\left(#1,#2\right)}
\newcommand{\isassembly}[1]{\mathtt{isAssembly}\left(#1\right)}
\newcommand{\disinhibitarea}[1]{\mathtt{disinhibitArea}\left(#1\right)}
\newcommand{\inhibitarea}[1]{\mathtt{inhibitArea}\left(#1\right)}
\newcommand{\disinhibitfiber}[1]{\mathtt{disinhibitFiber}\left(#1\right)}
\newcommand{\inhibitfiber}[1]{\mathtt{inhibitFiber}\left(#1\right)}
\newcommand{\activateblock}[1]{\mathtt{activateBlock}\left(#1\right)}

\newcommand{\workzone}[1]{\mbox{\textsc{Node}$_{#1}$}\xspace}
\newcommand{\blocks}{\textsc{Blocks}\xspace}
\newcommand{\headzone}{\textsc{Head}\xspace}

\newcommand{\parseralg}[1]{\textsc{Parser}\left (#1\right)}

\usepackage{xcolor}
\usepackage{blindtext}
 
\usepackage[linesnumbered,lined,boxruled,commentsnumbered]{algorithm2e}
\usepackage{array}
\usepackage{tikz}
\usetikzlibrary{shapes.geometric,positioning}
\usetikzlibrary{external}

\begin{document}

\maketitle              

\begin{abstract}
\begin{quote}
	We revisit the planning problem in the blocks world, and we implement a known heuristic for this task.  Importantly, our implementation is biologically plausible, in the sense that it is carried out exclusively through the spiking of neurons.  Even though much has been accomplished in the blocks world over the past five decades, we believe that this is the first algorithm of its kind.   The input is a sequence of symbols encoding an initial set of block stacks as well as a target set, and the output is a sequence of motion commands such as ``put the top block in stack 1 on the table''.  The program is written in the Assembly Calculus, a recently proposed computational framework meant to model computation in the brain by bridging the gap between neural activity and cognitive function.  Its elementary objects are assemblies of neurons (stable sets of neurons whose simultaneous firing signifies that the subject is thinking of an object, concept, word, etc.), its commands include project and merge, and its execution model is based on widely accepted tenets of neuroscience.  A program in this framework essentially sets up a dynamical system of neurons and synapses that eventually, with high probability, accomplishes the task.  The purpose of this work is to establish empirically that reasonably large programs in the Assembly Calculus can execute correctly and reliably; and that rather realistic --- if idealized --- higher cognitive functions, such as planning in the blocks world, can be implemented successfully by such programs.
\end{quote}
\end{abstract}

\section{Introduction}\label{sec:intro}
How does intelligence happen?   How can reasoning, problem-solving, decision-making, planning, empathy, language, art be achieved through the activity of neurons and synapses?  Despite tremendous advances over the past decades in our understanding of neural mechanisms --- increasingly assisted and propelled by machine learning --- we are still very far from answering the overarching question: {\em how does the brain beget the mind?}  The difficulty lies in the huge gap of scale and methodology between Experimental Neuroscience and Cognitive Science.  This frustration was articulated in a most eloquent way by Nobel laureate Richard Axel, who declared in a 2018 interview \cite{Axel2018}: {\em ``We do not have a logic for the transformation of neural activity to thought and action. I consider discerning [this logic] as the most important future direction in Neuroscience''.}

The \emph{Assembly Calculus} (AC) is a recently proposed formal computational system  \cite{papadimitriou2020}. As far as we know, it is the only computational system in the literature whose explicit purpose is to bridge through computation the gap between neurons and intelligence --- that is to say, to function as  Axel's logic.  The basic data item of the AC is the {\em assembly of neurons,} a large stable set of neurons believed to represent an idea, object, word, etc., while its operations (project, associate, merge, etc.) create and manipulate assemblies in response to stimuli and other brain events.  Importantly, these operations can be provably simulated through the activity of stylized neurons and synapses.  All said, the AC is a Turing complete computational system founded firmly on the basic principles of Neuroscience. In the next section, we provide a comprehensive introduction to the AC; however, the interested reader may want to read \cite{papadimitriou2020}.

So, is the AC the bridging ``logic'' sought by Axel?  One avenue for pursuing this important question is to demonstrate empirically that reasonably complex cognitive phenomena can be formulated and implemented in the AC framework. Indeed, in the original paper \cite{papadimitriou2020} it was argued that aspects of language generation can be handled by the operations of the AC, while in a very recent paper \cite{mitropolsky_biologically_2021}, a Parser implemented in the AC was demonstrated to analyze syntactically reasonably complex sentences of English, and it was argued that it can be generalized to more complex features as well as  other natural languages.

{\em Our contribution in this paper is to demonstrate that a program in the AC is capable of implementing reasonably sophisticated stylized planning strategies -- in particular, heuristics for solving tasks in the blocks world} \cite{gupta1991,slaney2001}.
A \emph{blocks-world configuration} is defined by a set of {\em stacks,} where a stack is a sequence of unique {\em blocks,} each sitting on top of the previous one.  A stack of size one is just a {\em block sitting on the table} (see e.g. Fig. \ref{fig:bwexamples}-A). 
A configuration can be manipulated by moving a block from the top of a stack (or from the table) to the top of another stack (or to the table).  A {\em task} in the blocks world is the following:  Given a
starting configuration $C_{init}$ and a goal configuration $C_{goal}$, 
find a sequence of actions which transforms $C_{init}$ into $C_{goal}$.
It was shown in \cite{gupta1992} that solving a task in the blocks world with the smallest number of actions is NP-Complete, and it was observed that the following provides a simple 2-approximation strategy: 
{\em Move to the table all blocks that are not in their final positions, and then move these blocks one by one to their final positions.}

Here we implement this strategy in the AC.  From the exposition of this implementation and demonstration --- which happens to employ representations and structures of a different style from those needed for language tasks \cite{papadimitriou2020,mitropolsky_biologically_2021} --- we believe that it will become clear that more complicated heuristics for solving related tasks can be similarly implemented in the AC. 

In fact, the kind of representations needed for planning, involving long ``chains'' of assemblies linked through strong synaptic connections, reveals a limitation of the AC which was not apparent before: we find empirically that there are limits --- depending on the parameters of the execution model, such as the number of excitatory neurons per brain area, synaptic density, synaptic plasticity, and assembly size --- on the length of such chains that can be implemented reliably.  As chaining is also used in the Turing machine simulation demonstrating the completeness of the AC \cite{papadimitriou2020}, such limitations are significant because they bound from above the space complexity --- and therefore the parallel time complexity --- of AC computations.  We briefly discuss and quantify this issue in the experimental validation section.

\subsection{Related Work}\label{sec:related}

Terry Winograd introduced the blocks world half a century ago as the context for his language understanding system SHRDLU \cite{winograd1971}, but since then {blocks-world planning}  has been widely investigated, primarily because such tasks appear to
capture several of the difficulties posed to planning systems \cite{gupta1991,gupta1992}.  There has been extensive work in AI on blocks world problems, including recently on leveraging ANNs for solving them, and learning to solve them from examples (e.g., the Neural Logic Machines of \cite{dong2018neural}, or Neural Turing Machines, which are used for related problem-solving tasks \cite{graves2014neural}).  

Bridging the gap between low-level models of neural activity in the brain and high-level symbolic systems modelling cognitive processes is a fundamental open problem in artificial intelligence and neuroscience at large \cite{doursat_bridging_2013,aaaiChady99}. 
Several computational cognitive-science papers address the problem of solving (or learning to solve) block-worlds tasks in higher-level computational models of cognition, such as ACT-R or SOAR (see for instance \cite{actr-soar-kennedy, kurup-thesis,Panov2017}).  In contrast to the present paper, however, these works utilize high-level languages and  data structures for the programming of these systems, without providing a link, as we do, to the behavior of stylized neurons and synapses, in an effort to remain as faithful as possible to the ways animal brains would solve these tasks. 

Less related to our problem is the  literature on block stacking (see, for example,~\cite{Hayashi2007,Tian2020}). These papers the focus on the ability of humans and chimpanzees to place a block on top of an existing tower without toppling it.

Finally, it is worth mentioning some previous works on solving planning tasks through spiking neural networks, such as \cite{rueckert2016,basanisi2020}, in which the attention is more focused on learning world models.

A spiking neural network framework not unlike ours is Nengo \cite{bekolay2014}. One important difference is that our framework focuses on the known behavior called assemblies which enable higher levels of abstraction such as the AC, and carrying out far more advanced tasks such as in \cite{mitropolsky_biologically_2021} and the present paper.

\section{The Assembly Calculus}\label{sec:ACintro}

The Assembly Calculus (AC) \cite{papadimitriou2020} is a computational system for modeling a dynamical system of firing neurons. In this system, there is a finite number of \textit{areas}, each containing $n$ neurons. The neurons of an area form a random Erd\H{o}s-R\'enyi directed graph $G_{n,p}$, where $p$ is the probability that two neurons of the area are connected. Moreover, certain ordered pairs of areas are  connected one to another through an Erd\H{o}s-R\'enyi directed bipartite graph $G_{n,p}$.  The directed connections between areas are called {\em fibers}.

In the AC, neurons in an area $A$ fire in discrete time steps, and are subject to stylized forms of {\em inhibition} and {\em plasticity}. For what concerns inhibition, at any time step, we assume only $k_A$ of the $n$ neurons fire, that is, the ones that previously received the highest total input from all other areas --- these $k_A$ neurons are sometimes called the {\em winners}. Plasticity is modelled by assuming that, if, at a given time step, neuron $x$ fires and, at the next time step, an out-neighbor neuron $y$ of $x$ fires, then the weight of the synapse from $x$ to $y$ (which is $1$ at the beginning) is multiplied by $(1+\beta_A)$, where $\beta_A>0$. 
In the original definition of the AC, a process of {\em homeostasis} was also modelled through a periodic renormalization, at a different time scale, of the synaptic weights, in order to avoid the generation of huge weights.  Such process is of course part of any realistic brain system, also providing a mechanism for {\em forgetting.} We will not implement here this feature of the model.

Lastly, yet importantly, the AC allows {\em inhibiting} and {\em disinhibiting} areas and fibers at different time steps. The exact mechanism through which areas and fibers are (dis)-inhibited may vary; in a recent paper modeling syntactic processing using the AC, \cite{mitropolsky_biologically_2021} model specific neurons as having (dis)-inhibitory effects on areas or fibers. In this work, (dis)-inhibition is always determined by which areas and fibers fired at the previous time step.

The most important emergent object in the AC is the {\em assembly}, that is, a stable set of $k_A$ highly interconnected neurons in an area $A$. It is emergent in the sense that assemblies are not a primitive of the model; instead, they are formed through its more basic operations. Assemblies are by now well known and widely studied in neuroscience, and are thought to represent concepts, ideas, objects, words, etc., and are increasingly believed in recent years to play a central role in cognitive processes \cite{buzsaki_neural_2010}, often called ``the alphabet of the brain'' \cite{buzsaki_brain_2021}. In terms of classical thinking in AI, one could think of assemblies as the boundary in the brain between sub-symbolic and symbolic computation.

The AC makes possible to perform certain {\em operations} with assemblies, described next --- in fact, it is through these operations that assemblies are created, in a way that guarantees high connectivity. In \cite{papadimitriou2020}, the authors demonstrate, both mathematically and through simulation, that these operations are ``possible" in the sense that they can be stably performed with high probability in the dynamical system of neurons outlined in the previous paragraphs.   
In this paper, we mostly make use of one of these operations: {\em projection} of an assembly in an area into another assembly in another area. 

Let us assume that an assembly $x$ of $k_A$ neurons of the area $A$ has just fired into an area $B$ (presumably through a disinhibited fiber going from $A$ to $B$), and assume that $B$ was quiescent at that time (no neurons were firing).  This will result in a set $w_1$ of $k_B$ neurons (the winners) firing at the next time step. Next, the neurons in $B$ will receive inputs not only from the $k_A$ neurons of the assembly in $A$, which will continue to fire, but also from the neurons in $w_1$ through recurrent connections within $B$: this will result in a set $w_2$ of $k_B$ neurons, (the new winners) firing at the next time step, and so on. It has been proved that, under appropriate values of the parameters $n, k_A, k_B, \beta,$ and $p$, this process converges with high probability to an assembly $y$ of $k_B$ neurons in $B$, which is called the projection of $x$ into $B$ and can be thought as a copy of $x$ in $B$ such that, from now on, $y$ will fire every time $x$ fires.

For a complete description of the AC the reader is referred to \cite{papadimitriou2020}, where in addition to stability of various assembly operations, it is also proved that, under certain assumptions, this computational system is capable of performing arbitrary computations as long as the space required does not exceed $n\over k_A$ (under much milder assumptions, $\sqrt{n\over k_A}$).  

In this paper, similarly to the Parser of \cite{mitropolsky_biologically_2021}, our AC programs work by projecting between all pairs of disinhibited areas along disinhibited fibers at each time step. For brevity, this operation, i.e. a simultaneous set of projections between multiple areas, is called {\em strong projection.} 

\begin{table*}[t]
    \centering
\resizebox{\columnwidth}{!}{%
    \begin{tabular}{||l||l|l||}
\hline
\textbf{Operation} & \multicolumn{1}{c|}{\textbf{Input}} & \multicolumn{1}{c||}{\textbf{Semantics}}\\
\hline\hline
$\activateblock{b}$ & Block number $b$ & Makes the assembly of the block $b$ in the area \blocks fire\\
$\disinhibitarea{A}$ & Set $A$ of areas & Disinhibit all the areas in $A$\\
$\disinhibitfiber{P}$ & Set $P$ of pairs of areas & Disinhibit the fibers between any pair of areas in $P$\\
$\inhibitarea{A}$ & Set $A$ of areas & Inhibit all the areas in $A$\\
$\inhibitfiber{P}$ & Set $P$ of pairs of areas & Inhibit the fibers between any pair of areas in $P$\\
$\isassembly{a}$ & Area $a$ & Verify whether there is an active assembly in the area $a$\\
$\project{a_1}{a_2}$ & Areas $a_1$ and $a_2$ & Executes a projection of (the active assembly in) the area $a_1$ to the area $a_2$\\
$\projectstar$ &  & Executes a strong projection involving all the disinhibited areas and fibers\\
\hline
\end{tabular}     
}
\caption{The AC operations (primitive and non primitive) used in the paper.}
    \label{tbl:acoperations}
\end{table*}

Our AC programs are described with the operations in Table \ref{tbl:acoperations}. Inhibition and disinhibition are primitives of the AC system, whereas strong projection (tantamount to a set of simultaneous projections) is an emergent property of the AC's dynamical system. We use several other such ``emergent" operations, i.e., that are not primitives of the AC system, but can be stably implemented with its basic operations. For example, we will make use of an operation which allows us to verify whether in a specific area there exists a stable assembly (as the result of a projection). In Table~\ref{tbl:acoperations}, we summarize the operations (primitive and non primitive) of the AC system, that we will use in this paper. Note that the block activation operation is a special operation, which causes an assembly (in a special area {\sc Blocks}) corresponding to the named block to fire. 
 
\section{The Blocks World AC Program}
A blocks world (BW) configuration is a set of {\em stacks,} where each stack is a sequence of {\em blocks}, from top to bottom.  Each block is assumed to be a unique integer between 1 and $s$. Two BW configurations, the initial and the target configuration, constitute the input to the AC program (see Figure~\ref{fig:bwexamples}-A).  We shall at first concentrate on {\em configurations with a single stack} --- already a meaningful problem --- and we shall eventually graduate to multiple stacks (see subsection 3.5 below).  We next describe four AC programs: (a) a program that takes the input --- a sequence of integers  representing a stack --- and creates a list-like structure, in a set of brain areas and fibers, for representing the stack; (b) a program that removes the top block of a stack thus represented; (c) a program that adds a new block to the represented stack; and (d) a program for computing the {\em intersection} of two stacks represented this way, that is, the longest common suffix of the two sequences, read from bottom to top. 

All four programs work on a common set of brain areas connected with bi-directional fibers:  the area \blocks contains a fixed assembly for every possible block (these assemblies are special, in that each can be activated explicitly as the presentation of the corresponding number in the input). There are four other areas used in our AC programs: \headzone, \workzone{0}, \workzone{1}, and \workzone{2}. \headzone is connected to the \workzone{0} area via fibers, while each \workzone{} area is connected to \blocks, and to each other in the shape of a triangle: \workzone{0} is connected with \workzone{1}, which is connected with \workzone{2}, which is connected with \workzone{0} (see Figure~\ref{fig:parserareas}-B). All of these areas are standard brain areas of the AC system, containing $n$ randomly connected neurons of which at most $k$ fire at any time.

\subsection{The Parser}
\IncMargin{1em}
\begin{algorithm*}
    	\SetKwInOut{Input}{input}

	\Input{a stack $S$ of blocks $b_1, b_2, \dots, b_s$.}
	\BlankLine
	
	$\disinhibitarea{\left\{\blocks,\headzone,\workzone{0}\right\}}$;
	$\disinhibitfiber{\left\{\left(\headzone,\workzone{0}\right),\left(\workzone{0},\blocks\right)\right\}}$\;\label{line:firstblockstarts}
	$\activateblock{b_1}$;
	$\projectstar$\;
	$\inhibitarea{\left\{\headzone\right\}}$;
	$\inhibitfiber{\left\{\left(\headzone,\workzone{0}\right),\left(\workzone{0},\blocks\right)\right\}}$\;\label{line:firstblockends}
	\ForEach{$i$ with $2\leq i \leq s$}{
	    $p=(i-2)\bmod{3}$; $c=(i-1)\bmod{3}$\;\label{line:nextblockstarts}
	    $\disinhibitarea{\left\{\workzone{c}\right\}}$;
	    $\disinhibitfiber{\left\{\left(\workzone{p},\workzone{c}\right),\left(\workzone{c},\blocks\right)\right\}}$\;
		$\activateblock{b_i}$; $\projectstar$\;
	    $\inhibitarea{\left\{\workzone{p}\right\}}$;
	    $\inhibitfiber{\left\{\left(\workzone{p},\workzone{c}\right),\left(\workzone{c},\blocks\right)\right\}}$\;
		\label{line:nextblockends}
	}
    $\inhibitarea{\left\{\blocks,\workzone{(s-1)\bmod 3}\right\}}$;     \caption{$\parseralg{S}$}
    \label{alg:parser}
\end{algorithm*}
\DecMargin{1em}

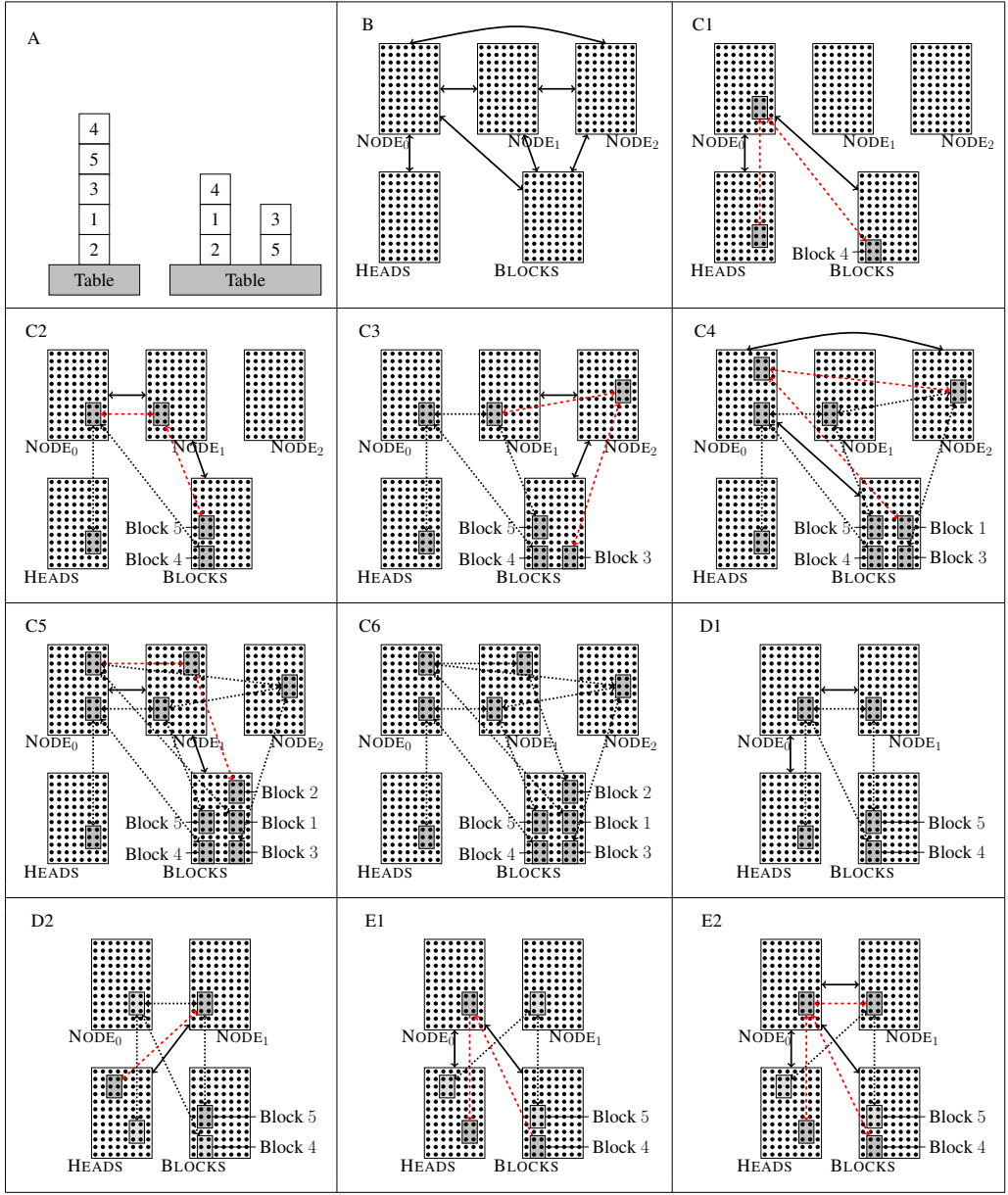
\begin{figure*}
    \centering
    {\renewcommand{\arraystretch}{2}}
    \resizebox{0.82\textwidth}{!}{\begin{tabular}{|c|c|c|}
        \hline
        & & \\
        \tikzstyle{block} = [rectangle, draw, minimum height=1cm, minimum width=1cm]
\tikzstyle{every node}=[font=\LARGE]
\begin{tikzpicture}[baseline]
	\node[block] at (0,5) {4};
	\node[block] at (0,4) {5};
	\node[block] at (0,3) {3};
	\node[block] at (0,2) {1};
	\node[block] at (0,1) {2};
	\node[rectangle, draw, minimum width=3cm, minimum height=1cm, fill=gray!50] at (0,0) {Table};
	\begin{scope}[xshift=3cm]
		\node[block] at (1,3) {4};
		\node[block] at (1,2) {1};
		\node[block] at (1,1) {2};
		\node[block] at (3,2) {3};
		\node[block] at (3,1) {5};
		\node[rectangle, draw, minimum width=5cm, minimum height=1cm, fill=gray!50] at (2,0) {Table};
	\end{scope}
	\node at (-2,8) {\LARGE A};
\end{tikzpicture} & \tikzstyle{area} = [rectangle,draw,minimum height=3cm,minimum width=2cm]
\tikzstyle{assembly} = [rectangle,draw,minimum height=0.75cm,minimum width=0.5cm]
\tikzstyle{every node}=[font=\LARGE]
\begin{tikzpicture}[x=0.25cm,y=0.25cm]
    \begin{scope}[xshift=1.5cm]
	\node[area] (B) at (3.5,5.5) {};
	\node at (0,-1.5) {\textsc{Blocks}};
	\foreach \x in {0,...,7}
	\foreach \y in {0,...,11}
	{\fill (\x,\y) circle (2pt);}
	\end{scope}
	\begin{scope}[xshift=-3.25cm,yshift=4.25cm]
	\node[area] (N1) at (3.5,5.5) {};
	\node at (0,-1.5) {\mbox{\textsc{Node}$_0$}};
	\foreach \x in {0,...,7}
	\foreach \y in {0,...,11}
	{\fill (\x,\y) circle (2pt);}
	\draw[<->,ultra thick] (B) -- (N1);
	\end{scope}
	\begin{scope}[xshift=0cm,yshift=4.25cm]
	\node[area] (N2) at (3.5,5.5) {};
	\node at (7,-1.5) {\mbox{\textsc{Node}$_1$}};
	\foreach \x in {0,...,7}
	\foreach \y in {0,...,11}
	{\fill (\x,\y) circle (2pt);}
	\draw[<->,ultra thick] (B) -- (N2);
	\draw[<->,ultra thick] (N1) -- (N2);
	\end{scope}
	\begin{scope}[xshift=3.25cm,yshift=4.25cm]
	\node[area] (N3) at (3.5,5.5) {};
	\node at (7,-1.5) {\mbox{\textsc{Node}$_2$}};
	\foreach \x in {0,...,7}
	\foreach \y in {0,...,11}
	{\fill (\x,\y) circle (2pt);}
	\draw[<->,ultra thick] (B) -- (N3);
	\draw[<->,ultra thick] (N2) -- (N3);
	\draw[bend right,<->,ultra thick] (N1.north) .. controls (-8,14.5) .. (N3.north);
	\end{scope}
	\begin{scope}[xshift=-3.25cm,yshift=0cm]
	\node[area] (H) at (3.5,5.5) {};
	\node at (0,-1.5) {\textsc{Heads}};
	\foreach \x in {0,...,7}
	\foreach \y in {0,...,11}
	{\fill (\x,\y) circle (2pt);}
	\draw[<->,ultra thick] (H) -- (N1);
	\end{scope}
	\node at (-15,31) {\LARGE B};
\end{tikzpicture} & \tikzstyle{area} = [rectangle,draw,minimum height=3cm,minimum width=2cm]
\tikzstyle{assembly} = [rectangle,draw,minimum height=0.75cm,minimum width=0.5cm]
\tikzstyle{every node}=[font=\LARGE]
\begin{tikzpicture}[x=0.25cm,y=0.25cm]
    \begin{scope}[xshift=1.5cm]
	\node[area] (B) at (3.5,5.5) {};
	\node at (0,-1.5) {\textsc{Blocks}};
	\node[assembly,fill=gray!50] (b1) at (1.5,1) {};
	\node[align=right] (b1t) at (-5.5,0.75) {Block $4$};
	\foreach \x in {0,...,7}
	\foreach \y in {0,...,11}
	{\fill (\x,\y) circle (2pt);}
	\draw (b1) -- (b1t);
	\end{scope}
	\begin{scope}[xshift=-3.25cm,yshift=4.25cm]
		\node[area] (N1) at (3.5,5.5) {};
		\node at (0,-1.5) {\mbox{\textsc{Node}$_0$}};
		\node[assembly,fill=gray!50] (b1n) at (5.5,3) {};
		\foreach \x in {0,...,7}
		\foreach \y in {0,...,11}
		{\fill (\x,\y) circle (2pt);}
		\draw[<->,ultra thick] (B) -- (N1);
		\draw[<->,ultra thick,color=red,dashed] (b1) -- (b1n);
	\end{scope}
	\begin{scope}[xshift=0cm,yshift=4.25cm]
		\node[area] (N2) at (3.5,5.5) {};
		\node at (7,-1.5) {\mbox{\textsc{Node}$_1$}};
		\foreach \x in {0,...,7}
		\foreach \y in {0,...,11}
		{\fill (\x,\y) circle (2pt);}
	\end{scope}
	\begin{scope}[xshift=3.25cm,yshift=4.25cm]
		\node[area] (N3) at (3.5,5.5) {};
		\node at (7,-1.5) {\mbox{\textsc{Node}$_2$}};
		\foreach \x in {0,...,7}
		\foreach \y in {0,...,11}
		{\fill (\x,\y) circle (2pt);}
	\end{scope}
	\begin{scope}[xshift=-3.25cm,yshift=0cm]
		\node[area] (H) at (3.5,5.5) {};
		\node at (0,-1.5) {\textsc{Heads}};
		\node[assembly,fill=gray!50] (b1h) at (5.5,3) {};
		\foreach \x in {0,...,7}
		\foreach \y in {0,...,11}
		{\fill (\x,\y) circle (2pt);}
		\draw[<->,ultra thick] (H) -- (N1);
		\draw[<->,ultra thick,color=red,dashed] (b1n) -- (b1h);
	\end{scope}
	\node at (-15,31) {\LARGE C1};
\end{tikzpicture}  \\
        & & \\
        \hline
        & & \\
        \tikzstyle{area} = [rectangle,draw,minimum height=3cm,minimum width=2cm]
\tikzstyle{assembly} = [rectangle,draw,minimum height=0.75cm,minimum width=0.5cm]
\tikzstyle{every node}=[font=\LARGE]
\begin{tikzpicture}[x=0.25cm,y=0.25cm]
    \begin{scope}[xshift=1.5cm]
	\node[area] (B) at (3.5,5.5) {};
	\node at (0,-1.5) {\textsc{Blocks}};
	\node[assembly,fill=gray!50] (b1) at (1.5,1) {};
	\node[align=right] (b1t) at (-5.5,0.75) {Block $4$};
	\node[assembly,fill=gray!50] (b2) at (1.5,5) {};
	\node[align=right] (b2t) at (-5.5,5) {Block $5$};
	\foreach \x in {0,...,7}
	\foreach \y in {0,...,11}
	{\fill (\x,\y) circle (2pt);}
	\draw (b1) -- (b1t);
	\draw (b2) -- (b2t);
	\end{scope}
	\begin{scope}[xshift=-3.25cm,yshift=4.25cm]
		\node[area] (N1) at (3.5,5.5) {};
		\node at (0,-1.5) {\mbox{\textsc{Node}$_0$}};
		\node[assembly,fill=gray!50] (b1n) at (5.5,3) {};
		\foreach \x in {0,...,7}
		\foreach \y in {0,...,11}
		{\fill (\x,\y) circle (2pt);}
		\draw[<->,ultra thick,dotted] (b1) -- (b1n);
	\end{scope}
	\begin{scope}[xshift=0cm,yshift=4.25cm]
		\node[area] (N2) at (3.5,5.5) {};
		\node at (6.5,-1.5) {\mbox{\textsc{Node}$_1$}};
		\node[assembly,fill=gray!50] (b2n) at (1.5,3) {};
		\foreach \x in {0,...,7}
		\foreach \y in {0,...,11}
		{\fill (\x,\y) circle (2pt);}
		\draw[<->,ultra thick] (B) -- (N2);
		\draw[<->,ultra thick] (N1) -- (N2);
		\draw[<->,ultra thick,color=red,dashed] (b1n) -- (b2n);
		\draw[<->,ultra thick,color=red,dashed] (b2) -- (b2n);
	\end{scope}
	\begin{scope}[xshift=3.25cm,yshift=4.25cm]
		\node[area] (N3) at (3.5,5.5) {};
		\node at (6.5,-1.5) {\mbox{\textsc{Node}$_2$}};
		\foreach \x in {0,...,7}
		\foreach \y in {0,...,11}
		{\fill (\x,\y) circle (2pt);}
	\end{scope}
	\begin{scope}[xshift=-3.25cm,yshift=0cm]
		\node[area] (H) at (3.5,5.5) {};
		\node at (0,-1.5) {\textsc{Heads}};
		\node[assembly,fill=gray!50] (b1h) at (5.5,3) {};
		\foreach \x in {0,...,7}
		\foreach \y in {0,...,11}
		{\fill (\x,\y) circle (2pt);}
		\draw[<->,ultra thick,dotted] (b1n) -- (b1h);
	\end{scope}
	\node at (-15,31) {\LARGE C2};
\end{tikzpicture} & \tikzstyle{area} = [rectangle,draw,minimum height=3cm,minimum width=2cm]
\tikzstyle{assembly} = [rectangle,draw,minimum height=0.75cm,minimum width=0.5cm]
\tikzstyle{every node}=[font=\LARGE]
\begin{tikzpicture}[x=0.25cm,y=0.25cm]
    \begin{scope}[xshift=1.5cm]
	\node[area] (B) at (3.5,5.5) {};
	\node at (0,-1.5) {\textsc{Blocks}};
	\node[assembly,fill=gray!50] (b1) at (1.5,1) {};
	\node[align=right] (b1t) at (-5.5,0.75) {Block $4$};
	\node[assembly,fill=gray!50] (b2) at (1.5,5) {};
	\node[align=right] (b2t) at (-5.5,5) {Block $5$};
	\node[assembly,fill=gray!50] (b3) at (5.5,1) {};
	\node[align=left] (b3t) at (12.5,1) {Block $3$};
	\foreach \x in {0,...,7}
	\foreach \y in {0,...,11}
	{\fill (\x,\y) circle (2pt);}
	\draw (b1) -- (b1t);
	\draw (b2) -- (b2t);
	\draw (b3) -- (b3t);
	\end{scope}
	\begin{scope}[xshift=-3.25cm,yshift=4.25cm]
		\node[area] (N1) at (3.5,5.5) {};
		\node at (0,-1.5) {\mbox{\textsc{Node}$_0$}};
		\node[assembly,fill=gray!50] (b1n) at (5.5,3) {};
		\foreach \x in {0,...,7}
		\foreach \y in {0,...,11}
		{\fill (\x,\y) circle (2pt);}
		\draw[<->,ultra thick,dotted] (b1) -- (b1n);
	\end{scope}
	\begin{scope}[xshift=0cm,yshift=4.25cm]
		\node[area] (N2) at (3.5,5.5) {};
		\node at (6.5,-1.5) {\mbox{\textsc{Node}$_1$}};
		\node[assembly,fill=gray!50] (b2n) at (1.5,3) {};
		\foreach \x in {0,...,7}
		\foreach \y in {0,...,11}
		{\fill (\x,\y) circle (2pt);}
		\draw[<->,ultra thick,dotted] (b1n) -- (b2n);
		\draw[<->,ultra thick,dotted] (b2) -- (b2n);
	\end{scope}
	\begin{scope}[xshift=3.25cm,yshift=4.25cm]
		\node[area] (N3) at (3.5,5.5) {};
		\node at (6.5,-1.5) {\mbox{\textsc{Node}$_2$}};
		\node[assembly,fill=gray!50] (b3n) at (5.5,6) {};
		\foreach \x in {0,...,7}
		\foreach \y in {0,...,11}
		{\fill (\x,\y) circle (2pt);}
		\draw[<->,ultra thick] (B) -- (N3);
		\draw[<->,ultra thick] (N2) -- (N3);
		\draw[<->,ultra thick,color=red,dashed] (b2n) -- (b3n);
		\draw[<->,ultra thick,color=red,dashed] (b3) -- (b3n);
	\end{scope}
	\begin{scope}[xshift=-3.25cm,yshift=0cm]
		\node[area] (H) at (3.5,5.5) {};
		\node at (0,-1.5) {\textsc{Heads}};
		\node[assembly,fill=gray!50] (b1h) at (5.5,3) {};
		\foreach \x in {0,...,7}
		\foreach \y in {0,...,11}
		{\fill (\x,\y) circle (2pt);}
		\draw[<->,ultra thick,dotted] (b1n) -- (b1h);
	\end{scope}
	\node at (-15,31) {\LARGE C3};
\end{tikzpicture} & \tikzstyle{area} = [rectangle,draw,minimum height=3cm,minimum width=2cm]
\tikzstyle{assembly} = [rectangle,draw,minimum height=0.75cm,minimum width=0.5cm]
\tikzstyle{every node}=[font=\LARGE]
\begin{tikzpicture}[x=0.25cm,y=0.25cm]
    \begin{scope}[xshift=1.5cm]
	\node[area] (B) at (3.5,5.5) {};
	\node at (0,-1.5) {\textsc{Blocks}};
	\node[assembly,fill=gray!50] (b1) at (1.5,1) {};
	\node[align=right] (b1t) at (-5.5,0.75) {Block $4$};
	\node[assembly,fill=gray!50] (b2) at (1.5,5) {};
	\node[align=right] (b2t) at (-5.5,5) {Block $5$};
	\node[assembly,fill=gray!50] (b3) at (5.5,1) {};
	\node[align=left] (b3t) at (12.5,1) {Block $3$};
	\node[assembly,fill=gray!50] (b4) at (5.5,5) {};
	\node[align=left] (b4t) at (12.5,5) {Block $1$};
	\foreach \x in {0,...,7}
	\foreach \y in {0,...,11}
	{\fill (\x,\y) circle (2pt);}
	\draw (b1) -- (b1t);
	\draw (b2) -- (b2t);
	\draw (b3) -- (b3t);
	\draw (b4) -- (b4t);
	\end{scope}
	\begin{scope}[xshift=-3.25cm,yshift=4.25cm]
		\node[area] (N1) at (3.5,5.5) {};
		\node at (0,-1.5) {\mbox{\textsc{Node}$_0$}};
		\node[assembly,fill=gray!50] (b1n) at (5.5,3) {};
		\node[assembly,fill=gray!50] (b4n) at (5.5,9) {};
		\foreach \x in {0,...,7}
		\foreach \y in {0,...,11}
		{\fill (\x,\y) circle (2pt);}
		\draw[<->,ultra thick] (B) -- (N1);
		\draw[<->,ultra thick,dotted] (b1) -- (b1n);
		\draw[<->,ultra thick,color=red,dashed] (b4) -- (b4n);
	\end{scope}
	\begin{scope}[xshift=0cm,yshift=4.25cm]
		\node[area] (N2) at (3.5,5.5) {};
		\node at (6.5,-1.5) {\mbox{\textsc{Node}$_1$}};
		\node[assembly,fill=gray!50] (b2n) at (1.5,3) {};
		\foreach \x in {0,...,7}
		\foreach \y in {0,...,11}
		{\fill (\x,\y) circle (2pt);}
		\draw[<->,ultra thick,dotted] (b1n) -- (b2n);
		\draw[<->,ultra thick,dotted] (b2) -- (b2n);
	\end{scope}
	\begin{scope}[xshift=3.25cm,yshift=4.25cm]
		\node[area] (N3) at (3.5,5.5) {};
		\node at (6.5,-1.5) {\mbox{\textsc{Node}$_2$}};
		\node[assembly,fill=gray!50] (b3n) at (5.5,6) {};
		\foreach \x in {0,...,7}
		\foreach \y in {0,...,11}
		{\fill (\x,\y) circle (2pt);}
		\draw[bend right,<->,ultra thick] (N1.north) .. controls (-8,14.5) .. (N3.north);
		\draw[<->,ultra thick,dotted] (b2n) -- (b3n);
		\draw[<->,ultra thick,dotted] (b3) -- (b3n);
		\draw[<->,ultra thick,color=red,dashed] (b3n) -- (b4n);
	\end{scope}
	\begin{scope}[xshift=-3.25cm,yshift=0cm]
		\node[area] (H) at (3.5,5.5) {};
		\node at (0,-1.5) {\textsc{Heads}};
		\node[assembly,fill=gray!50] (b1h) at (5.5,3) {};
		\foreach \x in {0,...,7}
		\foreach \y in {0,...,11}
		{\fill (\x,\y) circle (2pt);}
		\draw[<->,ultra thick,dotted] (b1n) -- (b1h);
	\end{scope}
	\node at (-15,31) {\LARGE C4};
\end{tikzpicture}  \\
        & & \\
        \hline
        & & \\
        \tikzstyle{area} = [rectangle,draw,minimum height=3cm,minimum width=2cm]
\tikzstyle{assembly} = [rectangle,draw,minimum height=0.75cm,minimum width=0.5cm]
\tikzstyle{every node}=[font=\LARGE]
\begin{tikzpicture}[x=0.25cm,y=0.25cm]
    \begin{scope}[xshift=1.5cm]
	\node[area] (B) at (3.5,5.5) {};
	\node at (0,-1.5) {\textsc{Blocks}};
	\node[assembly,fill=gray!50] (b1) at (1.5,1) {};
	\node[align=right] (b1t) at (-5.5,0.75) {Block $4$};
	\node[assembly,fill=gray!50] (b2) at (1.5,5) {};
	\node[align=right] (b2t) at (-5.5,5) {Block $5$};
	\node[assembly,fill=gray!50] (b3) at (5.5,1) {};
	\node[align=left] (b3t) at (12.5,1) {Block $3$};
	\node[assembly,fill=gray!50] (b4) at (5.5,5) {};
	\node[align=left] (b4t) at (12.5,5) {Block $1$};
	\node[assembly,fill=gray!50] (b5) at (5.5,9) {};
	\node[align=left] (b5t) at (12.5,9) {Block $2$};
	\foreach \x in {0,...,7}
	\foreach \y in {0,...,11}
	{\fill (\x,\y) circle (2pt);}
	\draw (b1) -- (b1t);
	\draw (b2) -- (b2t);
	\draw (b3) -- (b3t);
	\draw (b4) -- (b4t);
	\draw (b5) -- (b5t);
	\end{scope}
	\begin{scope}[xshift=-3.25cm,yshift=4.25cm]
		\node[area] (N1) at (3.5,5.5) {};
		\node at (0,-1.5) {\mbox{\textsc{Node}$_0$}};
		\node[assembly,fill=gray!50] (b1n) at (5.5,3) {};
		\node[assembly,fill=gray!50] (b4n) at (5.5,9) {};
		\foreach \x in {0,...,7}
		\foreach \y in {0,...,11}
		{\fill (\x,\y) circle (2pt);}
		\draw[<->,ultra thick,dotted] (b1) -- (b1n);
		\draw[<->,ultra thick,dotted] (b4) -- (b4n);
	\end{scope}
	\begin{scope}[xshift=0cm,yshift=4.25cm]
		\node[area] (N2) at (3.5,5.5) {};
		\node at (6.5,-1.5) {\mbox{\textsc{Node}$_1$}};
		\node[assembly,fill=gray!50] (b2n) at (1.5,3) {};
		\node[assembly,fill=gray!50] (b5n) at (5.5,9) {};
		\foreach \x in {0,...,7}
		\foreach \y in {0,...,11}
		{\fill (\x,\y) circle (2pt);}
		\draw[<->,ultra thick] (B) -- (N2);
		\draw[<->,ultra thick] (N1) -- (N2);
		\draw[<->,ultra thick,dotted] (b1n) -- (b2n);
		\draw[<->,ultra thick,dotted] (b2) -- (b2n);
		\draw[<->,ultra thick,color=red,dashed] (b5) -- (b5n);
		\draw[<->,ultra thick,color=red,dashed] (b4n) -- (b5n);
	\end{scope}
	\begin{scope}[xshift=3.25cm,yshift=4.25cm]
		\node[area] (N3) at (3.5,5.5) {};
		\node at (6.5,-1.5) {\mbox{\textsc{Node}$_2$}};
		\node[assembly,fill=gray!50] (b3n) at (5.5,6) {};
		\foreach \x in {0,...,7}
		\foreach \y in {0,...,11}
		{\fill (\x,\y) circle (2pt);}
		\draw[<->,ultra thick,dotted] (b2n) -- (b3n);
		\draw[<->,ultra thick,dotted] (b3) -- (b3n);
		\draw[<->,ultra thick,dotted] (b3n) -- (b4n);
	\end{scope}
	\begin{scope}[xshift=-3.25cm,yshift=0cm]
		\node[area] (H) at (3.5,5.5) {};
		\node at (0,-1.5) {\textsc{Heads}};
		\node[assembly,fill=gray!50] (b1h) at (5.5,3) {};
		\foreach \x in {0,...,7}
		\foreach \y in {0,...,11}
		{\fill (\x,\y) circle (2pt);}
		\draw[<->,ultra thick,dotted] (b1n) -- (b1h);
	\end{scope}
	\node at (-15,31) {\LARGE C5};
\end{tikzpicture} & \tikzstyle{area} = [rectangle,draw,minimum height=3cm,minimum width=2cm]
\tikzstyle{assembly} = [rectangle,draw,minimum height=0.75cm,minimum width=0.5cm]
\tikzstyle{every node}=[font=\LARGE]
\begin{tikzpicture}[x=0.25cm,y=0.25cm]
    \begin{scope}[xshift=1.5cm]
	\node[area] (B) at (3.5,5.5) {};
	\node at (0,-1.5) {\textsc{Blocks}};
	\node[assembly,fill=gray!50] (b1) at (1.5,1) {};
	\node[align=right] (b1t) at (-5.5,0.75) {Block $4$};
	\node[assembly,fill=gray!50] (b2) at (1.5,5) {};
	\node[align=right] (b2t) at (-5.5,5) {Block $5$};
	\node[assembly,fill=gray!50] (b3) at (5.5,1) {};
	\node[align=left] (b3t) at (12.5,1) {Block $3$};
	\node[assembly,fill=gray!50] (b4) at (5.5,5) {};
	\node[align=left] (b4t) at (12.5,5) {Block $1$};
	\node[assembly,fill=gray!50] (b5) at (5.5,9) {};
	\node[align=left] (b5t) at (12.5,9) {Block $2$};
	\foreach \x in {0,...,7}
	\foreach \y in {0,...,11}
	{\fill (\x,\y) circle (2pt);}
	\draw (b1) -- (b1t);
	\draw (b2) -- (b2t);
	\draw (b3) -- (b3t);
	\draw (b4) -- (b4t);
	\draw (b5) -- (b5t);
	\end{scope}
	\begin{scope}[xshift=-3.25cm,yshift=4.25cm]
		\node[area] (N1) at (3.5,5.5) {};
		\node at (0,-1.5) {\mbox{\textsc{Node}$_0$}};
		\node[assembly,fill=gray!50] (b1n) at (5.5,3) {};
		\node[assembly,fill=gray!50] (b4n) at (5.5,9) {};
		\foreach \x in {0,...,7}
		\foreach \y in {0,...,11}
		{\fill (\x,\y) circle (2pt);}
		\draw[<->,ultra thick,dotted] (b1) -- (b1n);
		\draw[<->,ultra thick,dotted] (b4) -- (b4n);
	\end{scope}
	\begin{scope}[xshift=0cm,yshift=4.25cm]
		\node[area] (N2) at (3.5,5.5) {};
		\node at (6.5,-1.5) {\mbox{\textsc{Node}$_1$}};
		\node[assembly,fill=gray!50] (b2n) at (1.5,3) {};
		\node[assembly,fill=gray!50] (b5n) at (5.5,9) {};
		\foreach \x in {0,...,7}
		\foreach \y in {0,...,11}
		{\fill (\x,\y) circle (2pt);}
		\draw[<->,ultra thick,dotted] (b1n) -- (b2n);
		\draw[<->,ultra thick,dotted] (b2) -- (b2n);
		\draw[<->,ultra thick,dotted] (b5) -- (b5n);
		\draw[<->,ultra thick,dotted] (b4n) -- (b5n);
	\end{scope}
	\begin{scope}[xshift=3.25cm,yshift=4.25cm]
		\node[area] (N3) at (3.5,5.5) {};
		\node at (6.5,-1.5) {\mbox{\textsc{Node}$_2$}};
		\node[assembly,fill=gray!50] (b3n) at (5.5,6) {};
		\foreach \x in {0,...,7}
		\foreach \y in {0,...,11}
		{\fill (\x,\y) circle (2pt);}
		\draw[<->,ultra thick,dotted] (b2n) -- (b3n);
		\draw[<->,ultra thick,dotted] (b3) -- (b3n);
		\draw[<->,ultra thick,dotted] (b3n) -- (b4n);
	\end{scope}
	\begin{scope}[xshift=-3.25cm,yshift=0cm]
		\node[area] (H) at (3.5,5.5) {};
		\node at (0,-1.5) {\textsc{Heads}};
		\node[assembly,fill=gray!50] (b1h) at (5.5,3) {};
		\foreach \x in {0,...,7}
		\foreach \y in {0,...,11}
		{\fill (\x,\y) circle (2pt);}
		\draw[<->,ultra thick,dotted] (b1n) -- (b1h);
	\end{scope}
	\node at (-15,31) {\LARGE C6};
\end{tikzpicture} & \tikzstyle{area} = [rectangle,draw,minimum height=3cm,minimum width=2cm]
\tikzstyle{assembly} = [rectangle,draw,minimum height=0.75cm,minimum width=0.5cm]
\tikzstyle{every node}=[font=\LARGE]
\begin{tikzpicture}[x=0.25cm,y=0.25cm]
	\node[area] (B) at (3.5,5.5) {};
	\node at (0,-1.5) {\textsc{Blocks}};
	\node[assembly,fill=gray!50] (b1) at (1.5,1) {};
	\node[align=left] (b1t) at (12.5,1) {Block $4$};
	\node[assembly,fill=gray!50] (b2) at (1.5,5) {};
	\node[align=left] (b2t) at (12.5,5) {Block $5$};
	\foreach \x in {0,...,7}
	\foreach \y in {0,...,11}
	{\fill (\x,\y) circle (2pt);}
	\draw (b1) -- (b1t);
	\draw (b2) -- (b2t);
	\begin{scope}[xshift=-3.25cm,yshift=4.25cm]
		\node[area] (N1) at (3.5,5.5) {};
		\node at (0,-1.5) {\mbox{\textsc{Node}$_0$}};
		\node[assembly,fill=gray!50] (n1) at (5.5,3) {};
		\foreach \x in {0,...,7}
		\foreach \y in {0,...,11}
		{\fill (\x,\y) circle (2pt);}
		\draw[<->,ultra thick,dotted] (b1) -- (n1);
	\end{scope}
	\begin{scope}[xshift=0cm,yshift=4.25cm]
		\node[area] (N2) at (3.5,5.5) {};
		\node at (7,-1.5) {\mbox{\textsc{Node}$_1$}};
		\node[assembly,fill=gray!50] (n2) at (1.5,3) {};
		\foreach \x in {0,...,7}
		\foreach \y in {0,...,11}
		{\fill (\x,\y) circle (2pt);}
		\draw[<->,ultra thick] (N1) -- (N2);
		\draw[<->,ultra thick,dotted] (b2) -- (n2);
		\draw[<->,ultra thick,dotted] (n1) -- (n2);
	\end{scope}
	\begin{scope}[xshift=-3.25cm,yshift=0cm]
		\node[area] (H) at (3.5,5.5) {};
		\node at (0,-1.5) {\textsc{Heads}};
		\node[assembly,fill=gray!50] (h) at (5.5,3) {};
		\foreach \x in {0,...,7}
		\foreach \y in {0,...,11}
		{\fill (\x,\y) circle (2pt);}
		\draw[<->,ultra thick] (H) -- (N1);
		\draw[<->,ultra thick,dotted] (n1) -- (h);
	\end{scope}
	\node at (-20,31) {\LARGE D1};
\end{tikzpicture}  \\
        & & \\
        \hline
        & & \\
        \tikzstyle{area} = [rectangle,draw,minimum height=3cm,minimum width=2cm]
\tikzstyle{assembly} = [rectangle,draw,minimum height=0.75cm,minimum width=0.5cm]
\tikzstyle{every node}=[font=\LARGE]
\begin{tikzpicture}[x=0.25cm,y=0.25cm]
	\node[area] (B) at (3.5,5.5) {};
	\node at (0,-1.5) {\textsc{Blocks}};
	\node[assembly,fill=gray!25] (b1) at (1.5,1) {};
	\node[align=left] (b1t) at (12.5,1) {Block $4$};
	\node[assembly,fill=gray!50] (b2) at (1.5,5) {};
	\node[align=left] (b2t) at (12.5,5) {Block $5$};
	\foreach \x in {0,...,7}
	\foreach \y in {0,...,11}
	{\fill (\x,\y) circle (2pt);}
	\draw (b1) -- (b1t);
	\draw (b2) -- (b2t);
	\begin{scope}[xshift=-3.25cm,yshift=4.25cm]
		\node[area] (N1) at (3.5,5.5) {};
		\node at (0,-1.5) {\mbox{\textsc{Node}$_0$}};
		\node[assembly,fill=gray!25] (n1) at (5.5,3) {};
		\foreach \x in {0,...,7}
		\foreach \y in {0,...,11}
		{\fill (\x,\y) circle (2pt);}
		\draw[<->,ultra thick,dotted] (b1) -- (n1);
	\end{scope}
	\begin{scope}[xshift=0cm,yshift=4.25cm]
		\node[area] (N2) at (3.5,5.5) {};
		\node at (6.5,-1.5) {\mbox{\textsc{Node}$_1$}};
		\node[assembly,fill=gray!50] (n2) at (1.5,3) {};
		\foreach \x in {0,...,7}
		\foreach \y in {0,...,11}
		{\fill (\x,\y) circle (2pt);}
		\draw[<->,ultra thick,dotted] (b2) -- (n2);
		\draw[<->,ultra thick,dotted] (n1) -- (n2);
	\end{scope}
	\begin{scope}[xshift=-3.25cm,yshift=0cm]
		\node[area] (H) at (3.5,5.5) {};
		\node at (0,-1.5) {\textsc{Heads}};
		\node[assembly,fill=gray!25] (h) at (5.5,3) {};
		\node[assembly,fill=gray!50] (nh) at (2.5,9) {};
		\foreach \x in {0,...,7}
		\foreach \y in {0,...,11}
		{\fill (\x,\y) circle (2pt);}
		\draw[<->,ultra thick] (H) -- (N2);
		\draw[<->,ultra thick,dotted] (n1) -- (h);
		\draw[<->,ultra thick,color=red,dashed] (n2) -- (nh);
	\end{scope}
	\node at (-20,31) {\LARGE D2};
\end{tikzpicture} & \tikzstyle{area} = [rectangle,draw,minimum height=3cm,minimum width=2cm]
\tikzstyle{assembly} = [rectangle,draw,minimum height=0.75cm,minimum width=0.5cm]
\tikzstyle{every node}=[font=\LARGE]
\begin{tikzpicture}[x=0.25cm,y=0.25cm]
	\node[area] (B) at (3.5,5.5) {};
	\node at (0,-1.5) {\textsc{Blocks}};
	\node[assembly,fill=gray!50] (b1) at (1.5,1) {};
	\node[align=left] (b1t) at (12.5,1) {Block $4$};
	\node[assembly,fill=gray!25] (b2) at (1.5,5) {};
	\node[align=left] (b2t) at (12.5,5) {Block $5$};
	\foreach \x in {0,...,7}
	\foreach \y in {0,...,11}
	{\fill (\x,\y) circle (2pt);}
	\draw (b1) -- (b1t);
	\draw (b2) -- (b2t);
	\begin{scope}[xshift=-3.25cm,yshift=4.25cm]
		\node[area] (N1) at (3.5,5.5) {};
		\node at (0,-1.5) {\mbox{\textsc{Node}$_0$}};
		\node[assembly,fill=gray!50] (n1) at (5.5,3) {};
		\foreach \x in {0,...,7}
		\foreach \y in {0,...,11}
		{\fill (\x,\y) circle (2pt);}
		\draw[->,ultra thick,color=red,dashed] (b1) -- (n1);
		\draw[<->,ultra thick] (B) -- (N1);
	\end{scope}
	\begin{scope}[xshift=0cm,yshift=4.25cm]
		\node[area] (N2) at (3.5,5.5) {};
		\node at (6.5,-1.5) {\mbox{\textsc{Node}$_1$}};
		\node[assembly,fill=gray!25] (n2) at (1.5,3) {};
		\foreach \x in {0,...,7}
		\foreach \y in {0,...,11}
		{\fill (\x,\y) circle (2pt);}
		\draw[<->,ultra thick,dotted] (b2) -- (n2);
	\end{scope}
	\begin{scope}[xshift=-3.25cm,yshift=0cm]
		\node[area] (H) at (3.5,5.5) {};
		\node at (0,-1.5) {\textsc{Heads}};
		\node[assembly,fill=gray!50] (h) at (5.5,3) {};
		\node[assembly,fill=gray!25] (nh) at (2.5,9) {};
		\foreach \x in {0,...,7}
		\foreach \y in {0,...,11}
		{\fill (\x,\y) circle (2pt);}
		\draw[<->,ultra thick] (H) -- (N1);
		\draw[->,ultra thick,color=red,dashed] (n1) -- (h);
		\draw[<->,ultra thick,dotted] (n2) -- (nh);
	\end{scope}
	\node at (-20,31) {\LARGE E1};
\end{tikzpicture} & \tikzstyle{area} = [rectangle,draw,minimum height=3cm,minimum width=2cm]
\tikzstyle{assembly} = [rectangle,draw,minimum height=0.75cm,minimum width=0.5cm]
\tikzstyle{every node}=[font=\LARGE]
\tikzstyle{every node}=[font=\LARGE]
\begin{tikzpicture}[x=0.25cm,y=0.25cm]
	\node[area] (B) at (3.5,5.5) {};
	\node at (0,-1.5) {\textsc{Blocks}};
	\node[assembly,fill=gray!50] (b1) at (1.5,1) {};
	\node[align=left] (b1t) at (12.5,1) {Block $4$};
	\node[assembly,fill=gray!25] (b2) at (1.5,5) {};
	\node[align=left] (b2t) at (12.5,5) {Block $5$};
	\foreach \x in {0,...,7}
	\foreach \y in {0,...,11}
	{\fill (\x,\y) circle (2pt);}
	\draw (b1) -- (b1t);
	\draw (b2) -- (b2t);
	\begin{scope}[xshift=-3.25cm,yshift=4.25cm]
		\node[area] (N1) at (3.5,5.5) {};
		\node at (0,-1.5) {\mbox{\textsc{Node}$_0$}};
		\node[assembly,fill=gray!50] (n1) at (5.5,3) {};
		\foreach \x in {0,...,7}
		\foreach \y in {0,...,11}
		{\fill (\x,\y) circle (2pt);}
		\draw[<->,ultra thick,color=red,dashed] (b1) -- (n1);
		\draw[<->,ultra thick] (B) -- (N1);
	\end{scope}
	\begin{scope}[xshift=0cm,yshift=4.25cm]
		\node[area] (N2) at (3.5,5.5) {};
		\node at (6.5,-1.5) {\mbox{\textsc{Node}$_1$}};
		\node[assembly,fill=gray!50] (n2) at (1.5,3) {};
		\foreach \x in {0,...,7}
		\foreach \y in {0,...,11}
		{\fill (\x,\y) circle (2pt);}
		\draw[<->,ultra thick] (N1) -- (N2);
		\draw[<->,ultra thick,dotted] (b2) -- (n2);
		\draw[<->,ultra thick,color=red,dashed] (n1) -- (n2);
	\end{scope}
	\begin{scope}[xshift=-3.25cm,yshift=0cm]
		\node[area] (H) at (3.5,5.5) {};
		\node at (0,-1.5) {\textsc{Heads}};
		\node[assembly,fill=gray!50] (h) at (5.5,3) {};
		\node[assembly,fill=gray!25] (nh) at (2.5,9) {};
		\foreach \x in {0,...,7}
		\foreach \y in {0,...,11}
		{\fill (\x,\y) circle (2pt);}
		\draw[<->,ultra thick] (H) -- (N1);
		\draw[<->,ultra thick,color=red,dashed] (n1) -- (h);
		\draw[<->,ultra thick,dotted] (n2) -- (nh);
	\end{scope}
	\node at (-20,31) {\LARGE E2};
\end{tikzpicture}  \\
        & & \\
        \hline
    \end{tabular}
    }
    \caption{
\textbf{A.} Two BW configurations. In the rest of the figure, we consider the BW configuration shown on the left.
\textbf{B.} The five areas used by the parser AC program, along with the connections through fibers.
\textbf{C1-6.} The behavior of the parser AC program. The black solid lines denote the fibers of Figure~B which are disinhibited. The red dashed lines denote the newly created connections between assemblies in different areas, while the black dotted lines denote the connections previously created.
\textbf{D1-2.} The behavior of the AC program which removes the block from the top of a stack, with input the data structure resulting from the parser execution (only the areas involved in the remove operation are shown). The black solid lines denote the fibers which are disinhibited. The red dashed lines denote the newly created connections between assemblies in different areas, while the black dotted lines denote the already existing connections.
\textbf{E1-2.} The behavior of the AC program which put the block $4$ on top of the stack, above the block $5$. The black solid lines denote the fibers which are disinhibited. The red dashed lines denote the newly created connections (unidirectional and bidirectional) between assemblies in different areas, while the black dotted lines denote the already existing connections.
    }
    \label{fig:put}
    \label{fig:pop}
    \label{fig:isabove} 
    \label{fig:parserareas}
    \label{fig:bwexamples}
    \label{fig:parser}
\end{figure*}

The parser (see Algorithm~\ref{alg:parser}) processes each block in a stack sequentially, starting from the top. When it analyses the first block (see  lines~\ref{line:firstblockstarts}-\ref{line:firstblockends}), the three areas \blocks, \headzone, and \workzone{0}, and the fibers between \headzone and \workzone{0} and between \workzone{0} and \blocks are disinhibited. The block assembly is then activated and a strong projection is performed, thus creating a connection between the assembly in \blocks corresponding to the block and an assembly in \workzone{0}, and between this latter assembly and an assembly in \headzone (see the red dashed lines in Figure~\ref{fig:parser}-C1). Successively, the \headzone area and the fibers between \headzone and \workzone{0} and between \workzone{0} and \blocks are inhibited. For each other block in the stack (see  lines~\ref{line:nextblockstarts}-\ref{line:nextblockends}), the \workzone{} area next to the one (i.e., \workzone{i \bmod 3}) currently disinhibited (i.e., \workzone{i+1 \bmod 3}) is disinhibited, and the fibers between this \workzone{} area and the \blocks area and between the two \workzone{} areas are disinhibited. The next block assembly is then activated and a strong projection is performed, creating a connection between the assembly in \blocks and an assembly in the \workzone{} area just disinhibited, and between this latter assembly and the assembly previously activated in the previous \workzone{} area (see the red dashed lines in the figures~\ref{fig:parser}-C2,C3,C4). After this and before the next block, this latter \workzone{} area and the fibers between it and the \workzone{} area after it, and those between the \workzone{} area after it and the \blocks area, are inhibited.

The final data structure is a {\em chain} of assemblies starting from an assembly in \headzone and passing through assemblies in the \workzone{} areas (see Figure~\ref{fig:parser}-C6). Note that this chain can contain more than one assembly in the same \workzone{} area: for instance, in Figure~\ref{fig:parser}-C6, the chain contains two assemblies in \workzone{0} and \workzone{1}. Each assembly in the chain is also connected to the assembly in \blocks corresponding to a block in the stack. For instance, the sequence of such assemblies in Figure~\ref{fig:parser}-C6 corresponds to the sequence of blocks $4,5,3,1,2$, which is exactly the sequence of blocks in the stack from top to bottom (see the left part of Figure~\ref{fig:bwexamples}-A). Note that Algorithm~\ref{alg:parser} uses a constant number of brain areas (that is, five), independently of the number of blocks in the stack.

 \subsection{Removing the Top Block}

In order to implement in AC the algorithm which transforms an input stack of blocks into a target stack of blocks, we start by describing an AC program to remove a block from the top of a stack. This program uses the same areas and fibers of the parser described in the previous section (see Figure~\ref{fig:parserareas}-B), with the addition of fibers between \headzone with \workzone{1}, and \headzone with \workzone{2}. Intuitively, these fibers are needed to allow changing the head of the chain representing the current stack, without having to shift all the assemblies one position to the left.

The AC program, which ``removes'' the block from the top of the stack, uses the connections created by the parser in order to activate the assembly in the \workzone{1}, which is connected to the block just below the top block (that is block $5$ in Figure~\ref{fig:pop}-D1,D2). This is done by projecting from the \headzone into \workzone{0}, and projecting from \workzone{0} into the \workzone{1} (see Figure~\ref{fig:pop}-D1). Through strong projection, the program successively creates a new connection from the active assembly in the \workzone{1} area to a new assembly in the \headzone area (see the red dashed line in Figure~\ref{fig:pop}-D2).

Note that the connections between the light gray assemblies in Figure~\ref{fig:pop}-D2 are still active, but they will not be used in the future since the last active assembly in the \headzone area is now connected to the assembly in the \workzone{1} area. These connections, indeed, might later disappear because of a process of homeostasis, which can be modeled in the AC system through a sort of ``renormalization'' (as described in~\cite{papadimitriou2020}). In a certain sense, the system will slowly ``forget'' which block was on the top of the stack, before a removal operation.

The removal of the top block can be repeated as many times as the number of blocks in the stack. The only difference is that the activation of the assembly in \workzone{} corresponding to the block below the top one is done by projecting \headzone into the \workzone{} area corresponding to the top block, and then projecting from this \workzone{} area to the one following it (in modular arithmetic).

In order to maintain an updated representation of the blocks world configuration, we use four additional brain areas to store the chain of blocks which have been removed and that, hence, are currently on the table. This chain can be implemented in the AC system exactly the same way we did when parsing a stack of blocks. Then, when we want to read the current data structure stored in the AC system, we examine the stack of blocks represented in \headzone and the \workzone{} areas, as well as the chain of blocks on the table in the additional areas.
 \subsection{Putting a Block on Top of the Stack}

The second operation we need in order to implement a minimal planning algorithm for the blocks world problem is {\em putting} a block on top of the stack. The AC program, for this operation first projects the block from in \blocks into the \workzone{} area preceding (in modular arithmetic) the \workzone{} area currently connected to  \headzone, and then projects the newly created assembly into \headzone (see Figure~\ref{fig:put}-E1). Successively, the program executes a strong projection between the four areas in order to correctly connect them (see Figure~\ref{fig:put}-E2). Once again, an active connection between the \headzone area and a \workzone{} area will still exist after the execution of the AC program, but this connection will not be used in the future. 

 \subsection{Computing the Intersection of Two Stacks}

The pop and put operations described in the previous two sections are sufficient to implement a simple planning algorithm, which consists in moving all the blocks on the table (by using pop), and by then moving the blocks on the table on top of the stack (by using put) according to the target stack.

In order to improve this algorithm and execute the two-approximation algorithm described in the introduction, we need an AC program which implements a third operation, that is, finding the {\em intersection} of two stacks. This operation looks for the common sub-stack of the two stacks (starting from the bottom) and return the highest block in this sub-stack. Then only the blocks above this block have to be moved on the table and reassembled in the right order.  

In a nutshell, this can be achieved in AC by first reaching the bottom of the two stacks which have to be compared, and then proceeding upwards until we find two different blocks, or the end of one of the two stacks.

\subsection{Multiple Stacks}
So far in this exposition we have concerned ourselves with configurations consisting of one stack.  In our experiments (see the next section) we have implemented up to five stacks by employing a different set of four areas for each stack. This is a bit unsatisfactory, because it implies that the maximum number of stacks that can be handled by the brain is encoded in the brain architecture. There is a rather simple --- in principle --- way to achieve the same effect by re-using the same four areas; we have an initial implementation of this idea, which we intend to test in the future.  

With multiple stacks one has to solve the {\em matching problem:} identifying pairs of stacks in the input and output that must be transformed one to the other. Naively, this can be done by comparing all pairs of stacks, but this entails effort that is quadratic in the number of stacks. This latter strategy is the one currently employed in our experiments. In the future, we intend to test a more principled way, based on {\em hashing} the stacks into their bottom element, and attending to any collisions. 

\section{Experiments}
\label{sec:experiments}

A software system for programming in the AC, as well as implementations of the algorithms described in this paper, have been written in Julia~\cite{bezanson2017julia}.
We make use of the Java generator for BW configurations available at~\cite{Koeman2020}, based on \cite{slaney2001}. 
We ran experiments on over 100 blocks-world configurations, with up to five stacks and 10, 20, and 30 blocks. 
The algorithm worked correctly in every instance.  We have used various settings of the parameters $n,k,p,\beta$ -- a particularly good set of parameters is $n=10^6,k=50,p=0.1,\beta=0.1$. Interestingly, the algorithms do not work in all parameter settings, because of limits on the chaining operation (see the next discussion). 
The Julia source code can be found at \cite{jBrain}.

In general, the amount of rounds of strong project (parallel spikings of neurons) needed to carry out the BW tasks seems to be around 35 spikes per block processed (parse, popped, or pushed), which, assuming roughly 50 Hz spikes for excitatory neurons in the brain, is around 1.4 seconds per operation. \paragraph{Limits of the AC.} 
An unexpected finding of our simulations is that they are stable only under very specific parameter settings. The bottleneck of the planning algorithms is in parsing the chain of blocks, that is, memorizing the sequence of blocks so they can be read out reliably. In isolation we call this operation ``chaining''.

\begin{figure}
    \centering
    \begin{subfigure}[b]{.5\columnwidth}
        \includegraphics[width=\textwidth]{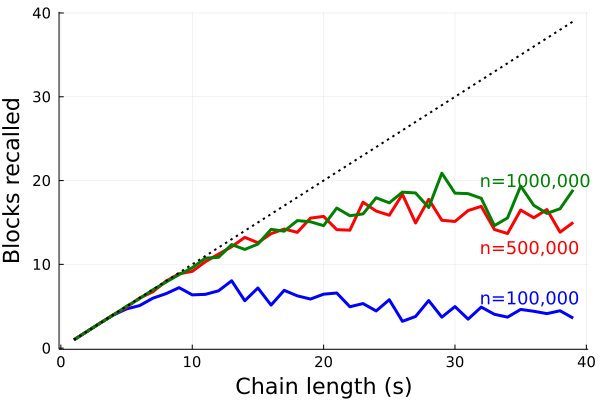}
        \caption{}
        \label{fig:vary-chain-length}
    \end{subfigure}
    \begin{subfigure}[b]{.5\columnwidth}
        \includegraphics[width=\textwidth]{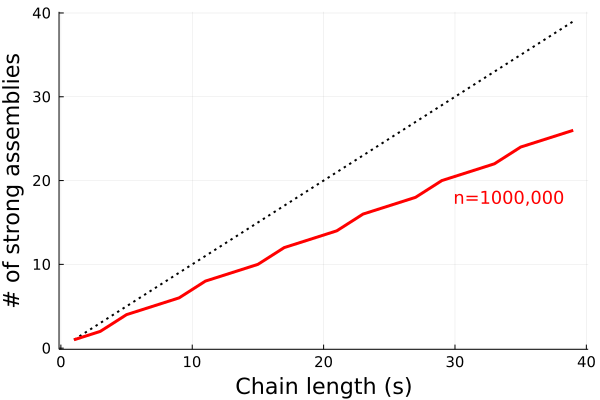}
        \caption{}
        \label{fig:strong-assemblies}
    \end{subfigure}
    \begin{subfigure}[b]{.5\columnwidth}
        \includegraphics[width=\textwidth]{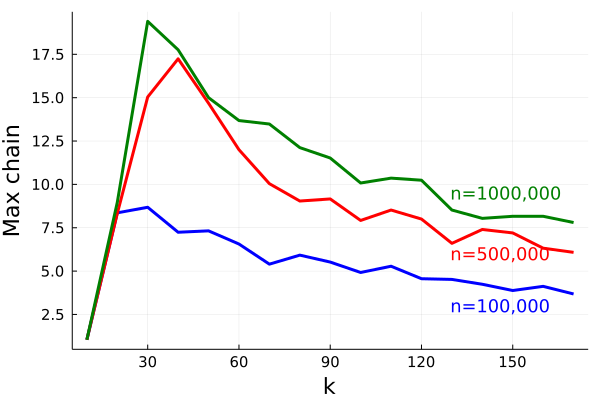}
        \caption{}
        \label{fig:vary-k}
    \end{subfigure}
    \caption{Experiments on the ``chaining" operation, the bottleneck of the AC  planning algorithm. (a) shows number of blocks correctly chained for various chain length; (b) shows number of ``strong" assemblies formed in chaining; (c) shows maximal chain length that is correctly parsed for varying $k$. (b) and (c) show averages over $50$ trials per parameter setting  (exact numbers, including sample standard deviation, are provided in the appendix). In these charts, $p=\beta=0.1$ was used, in (a) and (b) $k=50$.}
\end{figure}

The results in this section, which describe some properties and limits of chaining, can be viewed as theoretical properties of the AC. First, we find it is only possible to chain a rather limited number of blocks. For instance, even though with $n=10^6$ and $k=50$ there is, at least in theory, space for $10^6/50 = 200000$ non-overlapping assemblies, even with strong $p$ and $\beta$, we can only reliably chain up to $20$ blocks. This is illustrated in Figure \ref{fig:vary-chain-length}, which shows how many of $s$ blocks were successfully read out after chaining. Generally, for higher values of $n$ (and a higher $n:k$ ratio), longer portions of the chain tend to be correctly stored, but the operation is highly noisy: in some trials it will fail and then succeed for a {\em longer} chain. Indeed, unlike the assembly operations described in \cite{papadimitriou2020} (Project, Merge, and so on) which are either stable with overwhelming probability under appropriate parameters, or do not succeed if the parameters are not appropriately strong, chaining appears to push the computational power of the AC to its limits, and often succeeds or fails between repeated trials with the same parameters. 

One can also look at a related property: after chaining, how many of the assemblies in the $\workzone{i}$ areas during readout are ``strong'' in the sense that they pass \textsc{isAssembly()} with a high threshold (i.e. firing those $k$ neurons recursively results in the same set of $k$ winners)? Interestingly, this proportion, which is significantly less than the maximum of $s$, does not change significantly when we vary $n,p,\beta$-- there appears to be a natural proportion of strong assemblies formed during chaining (Figure \ref{fig:strong-assemblies}). 

Finally, in Figure \ref{fig:vary-k} we varied $k$ and found the maximally long chain that succeeded completely. These experiments again showed that for higher $n:k$ ratio, longer chains are possible, and that for each setting of $n$ there is a narrow window of optimal $k$ that allows for the longest chains-- above of this range, as we increase $k$ the maximum chain does not change, i.e. it appears to settle to some natural lower bound. A more thorough analysis of chaining is an important direction in AC theory, since such maneuvers could be subroutines in various cognitive processes (for instance, \cite{mitropolsky_biologically_2021} suggest using it for processing chains of identical parts of speech, such as multiple adjectives in a noun phrase). 

\section{Conclusions and Future Directions}\label{sec:conclusions}

The aim of this work is not so much to produce a performing system, but to demonstrate experimentally that reasonably large and complex programs in the assembly calculus can execute correctly and reliably, and in particular can implement in a natural manner planning strategies for solving instances of the blocks world problem. In fact, the implementation of these strategies is based on the realization of a \textit{list-like data structure} which makes use of a \textit{constant} number of brain regions. Confirming theoretical insights, we have experimentally found that the structure's reliability depends on the ratio between the number of neurons and the size of the assemblies in each region --- even though the dependency was a bit more constraining than we had expected.  The reasons and extent of this shortcoming must be the object of further investigation.

We have also shown how simple manipulations of the data structure (such as the top, pop, and append operations) can be realized by making use of a constant number of brain regions. These manipulations allowed us to implement planning strategies based on two basic kinds of moves, that is, moving the block from the top of a stack to the table, and putting a block from the table to the top of a stack. All our programs work for an \textit{arbitrary} number of blocks and a \textit{bounded} number of stacks --- while current work involves implementing a version with an arbitrary number of stacks.

After syntactic analysis in language and blocks world planning, what comes next as a compelling stylized cognitive function, which could be implemented in the AC?  There is work currently in submission dealing with {\em learning} though assemblies of neurons.  Two further realms of cognition come to mind, and they happen to be closely related:  {\em Reasoning,} as well as {\em planning and problem solving} in less specialized domains than BW. It would be interesting to figure out the most natural way for assemblies and their operations to carry out deductive tasks, and, even more ambitiously, to carry out planning in the context of logical and constraint-based formalisms of planning, see for example \cite{wilkins-book}.

\section*{Acknowledgements}
CHP's research was partially supported by NSF awards CCF1763970 and CCF1910700, by a research contract with Softbank, and a grant from CAIT. Furthermore, the authors are grateful to the OPAL infrastructure from Universit\'e C\^ote d'Azur for providing resources and support.

\bibliographystyle{abbrv}

\begin{thebibliography}{10}

\bibitem{Axel2018}
R.~Axel.
\newblock Q\&{A}.
\newblock {\em Neuron}, 99(6):1110--1112, Sept. 2018.
\newblock {doi: 10.1016/j.neuron.2018.09.003}.

\bibitem{basanisi2020}
R.~Basanisi, A.~Brovelli, E.~Cartoni, and G.~Baldassarre.
\newblock A generative spiking neural-network model of goal-directed behaviour
  and one-step planning.
\newblock {\em PLOS Computational Biology}, 16(12):1--32, 12 2020.

\bibitem{bekolay2014}
T.~Bekolay, J.~Bergstra, E.~Hunsberger, T.~DeWolf, T.~Stewart, D.~Rasmussen, X.~Choo, A.~Voelker, and C.~Eliasmith.
\newblock Nengo: a Python tool for building large-scale functional brain models.
\newblock {\em Frontiers in Neuroinformatics}, 7(48):1--13, 2014.

\bibitem{bezanson2017julia}
J.~Bezanson, A.~Edelman, S.~Karpinski, and V.~B. Shah.
\newblock Julia: A fresh approach to numerical computing.
\newblock {\em SIAM review}, 59(1):65--98, 2017.

\bibitem{buzsaki_neural_2010}
G.~Buzsáki.
\newblock Neural {Syntax}: {Cell} {Assemblies}, {Synapsembles}, and {Readers}.
\newblock {\em Neuron}, 68(3):362--385, Nov. 2010.

\bibitem{buzsaki_brain_2021}
G.~Buzsáki.
\newblock {\em The {Brain} from {Inside} {Out}}.
\newblock Oxford University Press, Oxford, reprint edition edition, Jan. 2021.

\bibitem{aaaiChady99}
M.~Chady.
\newblock Modelling higher cognitive functions with hebbian cell assemblies.
\newblock In J.~Hendler and D.~Subramanian, editors, {\em Proceedings of
  AAAI/IAAI 1999, July 18-22, 1999, Orlando, Florida, {USA}}, page 943. {AAAI}
  Press, 1999.

\bibitem{dong2018neural}
H.~Dong, J.~Mao, T.~Lin, C.~Wang, L.~Li, and D.~Zhou.
\newblock Neural logic machines.
\newblock In {\em International Conference on Learning Representations}, 2019.

\bibitem{doursat_bridging_2013}
R.~Doursat.
\newblock Bridging the {Mind}-{Brain} {Gap} by {Morphogenetic} "{Neuron}
  {Flocking}": {The} {Dynamic} {Self}-{Organization} of {Neural} {Activity}
  into {Mental} {Shapes}.
\newblock In {\em {AAAI} {Fall} {Symposia}}, 2013.

\bibitem{graves2014neural}
A.~Graves, G.~Wayne, and I.~Danihelka.
\newblock Neural turing machines, 2014.

\bibitem{gupta1991}
N.~Gupta and D.~S. Nau.
\newblock Complexity results for blocks-world planning.
\newblock In {\em Proceedings of the Ninth National Conference on Artificial
  Intelligence - Volume 2}, AAAI'91, page 629–633. AAAI Press, 1991.

\bibitem{gupta1992}
N.~Gupta and D.~S. Nau.
\newblock On the complexity of blocks-world planning.
\newblock {\em Artificial Intelligence}, 56(2):223--254, 1992.

\bibitem{Hayashi2007}
M.~Hayashi.
\newblock Stacking of blocks by chimpanzees: developmental processes and
  physical understanding.
\newblock {\em Animal Cognition}, 10:89--103, 2007.

\bibitem{jBrain}
jBrain.
\newblock \url{https://github.com/piluc/jBrain}, 12 2021.

\bibitem{actr-soar-kennedy}
W.~G. Kennedy and J.~G. Trafton.
\newblock Long-term symbolic learning in soar and act-r.
\newblock In {\em Proceedings of the Seventh International Conference on
  Cognitive Modeling}, page 166–171, 2006.

\bibitem{Koeman2020}
V.~Koeman.
\newblock The blocks world.
\newblock \url{https://github.com/eishub/blocksworld#readme}, 2020.
\newblock [Online; last access 08-September-2021].

\bibitem{kurup-thesis}
U.~Kurup.
\newblock {\em Design and use of a bimodal cognitive architecture for
  diagrammatic reasoning and cognitive modeling}.
\newblock {Ph.D.} diss., Graduate School of the Ohio State University, 2008.

\bibitem{mitropolsky_biologically_2021}
D.~Mitropolsky, M.~J. Collins, and C.~H. Papadimitriou.
\newblock A {Biologically} {Plausible} {Parser}.
\newblock In {\em Transactions of the {Association} for {Computational}
  {Linguistics}}, Aug. 2021.
\newblock arXiv: 2108.02189.

\bibitem{Panov2017}
A.~I. Panov.
\newblock Behavior planning of intelligent agent with sign world model.
\newblock {\em Biologically Inspired Cognitive Architectures}, 19:21--31, 2017.

\bibitem{papadimitriou2020}
C.~H. Papadimitriou, S.~S. Vempala, D.~Mitropolsky, M.~Collins, and W.~Maass.
\newblock Brain computation by assemblies of neurons.
\newblock {\em Proceedings of the National Academy of Sciences},
  117(25):14464--14472, 2020.

\bibitem{rueckert2016}
E.~Rueckert, D.~Kappel, D.~Tanneberg, D.~Pecevski, and J.~Peters.
\newblock Recurrent spiking networks solve planning tasks.
\newblock {\em Scientific Reports}, 6, 2016.

\bibitem{slaney2001}
J.~Slaney and S.~Thiébaux.
\newblock Blocks world revisited.
\newblock {\em Artificial Intelligence}, 125(1):119--153, 2001.

\bibitem{Tian2020}
M.~Tian, T.~Luo, and H.~Cheung.
\newblock The development and measurement of block construction in early
  childhood: A review.
\newblock {\em Journal of Psychoeducational Assessment}, 38(6):767--782, 2020.

\bibitem{wilkins-book}
D.~E. Wilkins.
\newblock {\em Practical Planning: Extending the Classical AI Planning
  Paradigm}.
\newblock Morgan Kaufmann Publishers Inc., San Francisco, CA, USA, 1988.

\bibitem{winograd1971}
T.~Winograd.
\newblock Procedures as a representation for data in a computer program for
  understanding natural language.
\newblock Technical report, Massachusetts Inst Of Tech Cambridge Project Mac,
  1971.

\end{thebibliography}

\appendix
\section{More Details on the Experiments}

The operation of parsing works in every possible instance we tried provided that the constraints shown in the subsection on the limits of AC are met. We tried parsing randomly generated BW configurations with 10, 20, and 30 blocks divided in multiple stacks, with the following parameters: $n = 4\times 10^6$ neurons, $p = 0.1, \beta = 0.1, k = 50$. The intersect operation needs two parsed stacks as input, and runs correctly each time the parsing operation of the stacks works correctly. Removing the top block (that is, the pop operation) and putting on top of a stack (that is, the put operation) run as well in the aforementioned settings. The used machine is a DELL laptop with an Intel(R) Core(TM) i7-8665U CPU @ 1.90GHz processor, 32GB Ram, running Fedora 33. The input BW configurations are specified in the \texttt{bw\_instances} folder at \cite{jBrain}.

Stricter constraints are needed for the whole planning operation. Since the previous operations have to be run many times and one after the other other, the graph representing the brain and its connectivity grows quite quickly. On a machine like the one described above, only BW configurations with at most 10 blocks are well handled (good parameters to test this are the same as for parsing). Otherwise, the program requires too much time in order to be completed, even with the same set of parameters. For the planning with 10 blocks, we used a machine Dell R940 quad-Xeon SP Gold 6148 @ 2.40GHz (80 cores) with 1024 GB of RAM.

In the case of BW configurations with 10 blocks, we have verified the correctness of the AC programs implementing the two planning algorithms (the one without and the one with the intersect operation) on 100 BW configurations randomly chosen, such that each stack has at most 7 blocks, to avoid chaining issues (these configurations are specified in the file \texttt{planning\_inputs.txt}). The execution on all these instances correctly run and finished in reasonable time.

Due to the discussion on the limits of AC (i.e., the limits on the maximum chain lengths), in the case of BW configurations between 20 and 30 blocks, we have limited ourselves to verify the correctness of the implementation of the basic operations used by the two algorithms, that is, the parser, the pop, the push, and the intersect operations (roughly 30 simulations). All runs were completed in reasonable time (few minutes - up to 20 in the case of more demanding operations) without errors. Also in this case, we used $n=4\times 10^6$ neurons for each brain area. We remark that the 20 and 30 blocks must be split among several stacks of maximum length up to 7, otherwise the parsing procedure may fail with the above number of neurons (these BW configurations are specified in the file \texttt{operation\_inputs.txt}).

The verification of the entire planning algorithms in the case of 20 and 30 blocks (even split among several stacks of up to 7 blocks each) requires more memory and time, due to the large computations needed to represent ``bigger'' brains. After running the planning algorithms in these cases, nevertheless, we observed that the initial actions performed by the brain were correct, which makes us believe that the algorithms would also correctly work in its entirety. 

\paragraph{Limits of the AC.}
Table \ref{table:chain_length_exp} shows the outcome of the experiments on chaining. In particular, it shows the mean number of blocks (over 50 runs) we can chain with $n = 10^5, 5 \times 10^5, 10^6$ neurons for each brain area, $k = 50$ (the number of neurons an assembly is composed of), and the standard deviation. With $10^6$ neurons, we can reliably chain up to $10,11$ blocks, but it's better if the number of blocks is less than 8. If we lower the number of neurons, less blocks can be reliably parsed.

Table \ref{table:maxchain_given_k} shows the outcome of the experiments on the ratio between $n$ and $k$ which can chain the higher number of blocks. The best reliability is obtained when $k$ increases together with $n$. For $n=10^5$, $20 \le k \le 30$ seems to be best. For $n = 5 \times 10^5 , 10^6$, $30\le k \le 40$ works better. The means and the standard deviations are obtained over 50 runs of the experiment.

In order to execute these experiments, the reader can execute the following terminal command:
\begin{verbatim}
	julia experiments/chaining_experiments.jl
\end{verbatim}

\begin{table*}[t]
\centering
\begin{tabular}{||l|l|l|l||l|l|l|l||l|l|l|l||}
\hline
\textbf{Neurons}     & \textbf{Blocks} & \textbf{Mean} & \textbf{Std}  & \textbf{Neurons}     & \textbf{Blocks} & \textbf{Mean} & \textbf{Std}  &  \textbf{Neurons}     & \textbf{Blocks} & \textbf{Mean} & \textbf{Std}    \\ \hline \hline
$10^5$ & 1             & 1    & 0    & $5 \cdot 10^5$ & 1             & 1     & 0     & $10^6$ & 1             & 1     & 0     \\ \hline
$10^5$ & 2             & 2    & 0    & $5 \cdot 10^5$ & 2             & 2     & 0     & $10^6$ & 2             & 2     & 0     \\ \hline
$10^5$ & 3             & 3    & 0    & $5 \cdot 10^5$ & 3             & 3     & 0     & $10^6$ & 3             & 3     & 0     \\ \hline
$10^5$ & 4             & 3,96 & 0,28 & $5 \cdot 10^5$ & 4             & 4     & 0     & $10^6$ & 4             & 3,96  & 0,28  \\ \hline
$10^5$ & 5             & 4,68 & 0,94 & $5 \cdot 10^5$ & 5             & 4,88  & 0,63  & $10^6$ & 5             & 5     & 0     \\ \hline
$10^5$ & 6             & 5,08 & 1,86 & $5 \cdot 10^5$ & 6             & 6     & 0     & $10^6$ & 6             & 5,92  & 0,57  \\ \hline
$10^5$ & 7             & 5,96 & 2,13 & $5 \cdot 10^5$ & 7             & 6,72  & 1,21  & $10^6$ & 7             & 6,96  & 0,28  \\ \hline
$10^5$ & 8             & 6,5  & 2,64 & $5 \cdot 10^5$ & 8             & 8     & 0     & $10^6$ & 8             & 7,88  & 0,63  \\ \hline
$10^5$ & 9             & 7,22 & 3,22 & $5 \cdot 10^5$ & 9             & 8,88  & 0,85  & $10^6$ & 9             & 8,78  & 1,3   \\ \hline
$10^5$ & 10            & 6,36 & 3,86 & $5 \cdot 10^5$ & 10            & 9,16  & 2,22  & $10^6$ & 10            & 9,6   & 1,81  \\ \hline
$10^5$ & 11            & 6,42 & 4,53 & $5 \cdot 10^5$ & 11            & 10,32 & 1,91  & $10^6$ & 11            & 10,74 & 1,38  \\ \hline
$10^5$ & 12            & 6,84 & 4,4  & $5 \cdot 10^5$ & 12            & 11,22 & 2,47  & $10^6$ & 12            & 10,84 & 2,57  \\ \hline
$10^5$ & 13            & 8,04 & 4,69 & $5 \cdot 10^5$ & 13            & 12,1  & 2,77  & $10^6$ & 13            & 12,36 & 2,45  \\ \hline
$10^5$ & 14            & 5,66 & 4,98 & $5 \cdot 10^5$ & 14            & 13,22 & 2,74  & $10^6$ & 14            & 11,78 & 4,26  \\ \hline
$10^5$ & 15            & 7,18 & 5,46 & $5 \cdot 10^5$ & 15            & 12,58 & 4,52  & $10^6$ & 15            & 12,42 & 4,51  \\ \hline
$10^5$ & 16            & 5,14 & 4,89 & $5 \cdot 10^5$ & 16            & 13,66 & 4,35  & $10^6$ & 16            & 14,18 & 3,77  \\ \hline
$10^5$ & 17            & 6,9  & 5,82 & $5 \cdot 10^5$ & 17            & 14,22 & 5,41  & $10^6$ & 17            & 13,96 & 5,03  \\ \hline
$10^5$ & 18            & 6,24 & 5,25 & $5 \cdot 10^5$ & 18            & 13,82 & 6,73  & $10^6$ & 18            & 15,22 & 4,99  \\ \hline
$10^5$ & 19            & 5,86 & 4,78 & $5 \cdot 10^5$ & 19            & 15,52 & 5,87  & $10^6$ & 19            & 15,08 & 6,19  \\ \hline
$10^5$ & 20            & 6,44 & 5,91 & $5 \cdot 10^5$ & 20            & 15,72 & 6,25  & $10^6$ & 20            & 14,62 & 6,66  \\ \hline
$10^5$ & 21            & 6,58 & 4,99 & $5 \cdot 10^5$ & 21            & 14,14 & 8,16  & $10^6$ & 21            & 16,72 & 6,87  \\ \hline
$10^5$ & 22            & 4,94 & 4,55 & $5 \cdot 10^5$ & 22            & 14,08 & 7,85  & $10^6$ & 22            & 15,8  & 7,69  \\ \hline
$10^5$ & 23            & 5,32 & 5,73 & $5 \cdot 10^5$ & 23            & 17,42 & 7,85  & $10^6$ & 23            & 16,02 & 8,04  \\ \hline
$10^5$ & 24            & 4,44 & 3,97 & $5 \cdot 10^5$ & 24            & 16,36 & 7,51  & $10^6$ & 24            & 17,94 & 7,88  \\ \hline
$10^5$ & 25            & 5,78 & 5,07 & $5 \cdot 10^5$ & 25            & 15,86 & 9,67  & $10^6$ & 25            & 17,34 & 9,2   \\ \hline
$10^5$ & 26            & 3,2  & 2,85 & $5 \cdot 10^5$ & 26            & 18,36 & 9     & $10^6$ & 26            & 18,62 & 9,24  \\ \hline
$10^5$ & 27            & 3,78 & 4,11 & $5 \cdot 10^5$ & 27            & 14,94 & 10,15 & $10^6$ & 27            & 18,52 & 9,36  \\ \hline
$10^5$ & 28            & 5,68 & 4,76 & $5 \cdot 10^5$ & 28            & 17,76 & 10,88 & $10^6$ & 28            & 16,78 & 9,76  \\ \hline
$10^5$ & 29            & 3,7  & 3,42 & $5 \cdot 10^5$ & 29            & 15,26 & 9,08  & $10^6$ & 29            & 20,88 & 10,21 \\ \hline
$10^5$ & 30            & 4,96 & 4,18 & $5 \cdot 10^5$ & 30            & 15,12 & 10,13 & $10^6$ & 30            & 18,5  & 11,14 \\ \hline
$10^5$ & 31            & 3,46 & 3,72 & $5 \cdot 10^5$ & 31            & 16,42 & 12    & $10^6$ & 31            & 18,44 & 11,79 \\ \hline
$10^5$ & 32            & 4,9  & 3,87 & $5 \cdot 10^5$ & 32            & 16,92 & 10,97 & $10^6$ & 32            & 17,88 & 11,87 \\ \hline
$10^5$ & 33            & 4,04 & 3,81 & $5 \cdot 10^5$ & 33            & 14,16 & 11,63 & $10^6$ & 33            & 14,62 & 11,38 \\ \hline
$10^5$ & 34            & 3,7  & 3,59 & $5 \cdot 10^5$ & 34            & 13,68 & 10,82 & $10^6$ & 34            & 15,54 & 12,04 \\ \hline
$10^5$ & 35            & 4,62 & 4,7  & $5 \cdot 10^5$ & 35            & 16,48 & 11,78 & $10^6$ & 35            & 19,34 & 11,9  \\ \hline
$10^5$ & 36            & 4,4  & 4,16 & $5 \cdot 10^5$ & 36            & 15,56 & 12,73 & $10^6$ & 36            & 17,04 & 12,71 \\ \hline
$10^5$ & 37            & 4,12 & 3,49 & $5 \cdot 10^5$ & 37            & 16,54 & 12,67 & $10^6$ & 37            & 16,08 & 12,29 \\ \hline
$10^5$ & 38            & 4,46 & 3,88 & $5 \cdot 10^5$ & 38            & 13,84 & 12,86 & $10^6$ & 38            & 16,64 & 13,46 \\ \hline
$10^5$ & 39            & 3,58 & 3,24 & $5 \cdot 10^5$ & 39            & 15    & 12,68 & $10^6$ & 39            & 18,9  & 13,67 \\ \hline
\end{tabular}
\caption{Table for the chain length experiment. We check if  $n \in \{10^5, 5\cdot 10^5, 10^6\}$ neurons for brain area are sufficient to parse a stack of a given number of blocks. Here, the mean number of correctly parsed blocks and its standard deviation are shown over 50 runs of the experiment.}
\label{table:chain_length_exp}
\end{table*}

\begin{table*}
\centering
\begin{tabular}{||l|l|l|l||l|l|l|l||l|l|l|l||}
\hline
\textbf{Neurons}     & $k$  & \textbf{Mean} & \textbf{Std}  & \textbf{Neurons}     & $k$  & \textbf{Mean} & \textbf{Std}  & \textbf{Neurons}     & $k$  & \textbf{Mean} & \textbf{Std}   \\ \hline \hline
$10^5$ & 10  & 1,04 & 0,28 & $5\cdot 10^5$ & 10  & 1,04  & 0,28 & $10^6$ & 10  & 1,04  & 0,28 \\ \hline
$10^5$ & 20  & 8,36 & 4,25 & $5\cdot 10^5$ & 20  & 8,32  & 5,22 & $10^6$ & 20  & 8,96  & 4,47 \\ \hline
$10^5$ & 30  & 8,68 & 3,03 & $5\cdot 10^5$ & 30  & 15,04 & 6,38 & $10^6$ & 30  & 19,4  & 6,93 \\ \hline
$10^5$ & 40  & 7,24 & 2,93 & $5\cdot 10^5$ & 40  & 17,24 & 5,68 & $10^6$ & 40  & 17,76 & 6,13 \\ \hline
$10^5$ & 50  & 7,32 & 2,47 & $5\cdot 10^5$ & 50  & 14,68 & 5,53 & $10^6$ & 50  & 15    & 5,51 \\ \hline
$10^5$ & 60  & 6,56 & 2,43 & $5\cdot 10^5$ & 60  & 12    & 4,15 & $10^6$ & 60  & 13,68 & 5,01 \\ \hline
$10^5$ & 70  & 5,4  & 2,06 & $5\cdot 10^5$ & 70  & 10,04 & 2,98 & $10^6$ & 70  & 13,48 & 4,92 \\ \hline
$10^5$ & 80  & 5,92 & 1,99 & $5\cdot 10^5$ & 80  & 9,04  & 3,86 & $10^6$ & 80  & 12,12 & 4,34 \\ \hline
$10^5$ & 90  & 5,52 & 1,97 & $5\cdot 10^5$ & 90  & 9,16  & 3,1  & $10^6$ & 90  & 11,52 & 3,87 \\ \hline
$10^5$ & 100 & 4,92 & 1,98 & $5\cdot 10^5$ & 100 & 7,92  & 2,83 & $10^6$ & 100 & 10,08 & 3,96 \\ \hline
$10^5$ & 110 & 5,28 & 1,57 & $5\cdot 10^5$ & 110 & 8,52  & 2,82 & $10^6$ & 110 & 10,36 & 3,73 \\ \hline
$10^5$ & 120 & 4,56 & 1,58 & $5\cdot 10^5$ & 120 & 8     & 2,92 & $10^6$ & 120 & 10,24 & 3,38 \\ \hline
$10^5$ & 130 & 4,52 & 1,64 & $5\cdot 10^5$ & 130 & 6,6   & 2,56 & $10^6$ & 130 & 8,52  & 4,05 \\ \hline
$10^5$ & 140 & 4,24 & 1,33 & $5\cdot 10^5$ & 140 & 7,4   & 2,71 & $10^6$ & 140 & 8,04  & 2,78 \\ \hline
$10^5$ & 150 & 3,88 & 1,15 & $5\cdot 10^5$ & 150 & 7,2   & 2,43 & $10^6$ & 150 & 8,16  & 2,97 \\ \hline
$10^5$ & 160 & 4,12 & 1,08 & $5\cdot 10^5$ & 160 & 6,32  & 2,12 & $10^6$ & 160 & 8,16  & 3,23 \\ \hline
$10^5$ & 170 & 3,68 & 1,19 & $5\cdot 10^5$ & 170 & 6,08  & 2,18 & $10^6$ & 170 & 7,8   & 3,21 \\ \hline
\end{tabular}
\caption{Table for the max chain length experiment. We check, for a given number of $k$ (the number of neurons an assembly is composed of), the maximum chain length the brain can correctly parse, with $n \in \{10^5, 5\cdot 10^5, 10^6\}$ neurons in each area of the brain. Here, the mean number of the maximum chain length correctly parsed blocks and its standard deviation are shown over 50 runs of the experiment.}
\label{table:maxchain_given_k}
\end{table*}
 \end{document}